\documentclass[lettersize,journal]{IEEEtran}
\usepackage{amsmath,amsfonts}
\usepackage{algorithmic}
\usepackage{algorithm}
\usepackage{array}
\usepackage[caption=false,font=normalsize,labelfont=sf,textfont=sf]{subfig}
\usepackage{balance}

\usepackage{stfloats}
\usepackage{url}
\usepackage{verbatim}
\usepackage{multirow}
\usepackage{gensymb}
\usepackage{textcomp}
\usepackage{booktabs}
\usepackage{soul}
\usepackage{hyperref}
\usepackage{colortbl}
\usepackage[x11names,dvipsnames,table]{xcolor}

\usepackage[normalem]{ulem}
\useunder{\uline}{\ul}{}
\usepackage{graphicx}
\usepackage{cite}
\begin{document}

% Command for review
\newcommand{\remove}[1]{\textcolor{red}{\st{#1}}}
\newcommand{\add}[1]{\textcolor{blue}{#1}}
\newcommand{\replace}[2]{\textcolor{red}{\st{#1}} \textcolor{blue}{#2}}

%\captionsetup[subfloat]{font=footnotesize, labelfont=sf, textfont=sf}
\captionsetup[subfloat]{%
            font=footnotesize,
            labelformat=parens,labelsep=space,
            listofformat=subparens}

\title{A recurrent CNN for online object detection on raw radar frames}

\author{Colin Decourt$^{1,2,3,4}$, Rufin VanRullen$^{1,2}$, Didier Salle$^{1,4}$,  Thomas Oberlin $^{1,3}$% <-this % stops a space
\thanks{$^{1}$Artificial and Natural Intelligence Toulouse Institute, Université de Toulouse, France
        }%
\thanks{$^{2}$CerCO, CNRS UMR5549, Toulouse
       }%
\thanks{$^{3}$ISAE-SUPAERO, Université de Toulouse, 10 Avenue Edouard Belin, Toulouse 31400, France
       }%
\thanks{$^{4}$NXP Semiconductors, Toulouse, France}%
\thanks{This work has been funded by the Institute for Artificial and Natural Intelligence Toulouse (ANITI) under grant agreement ANR-19-PI3A-0004.}
\thanks{The code and the pre-trained weights is available at \url{https://github.com/colindecourt/record}}
}

%\thanks{This paper was produced by the IEEE Publication Technology Group. They are in Piscataway, NJ.}% <-this % stops a space
%\thanks{Manuscript received April 19, 2021; revised August 16, 2021.}}



\maketitle

\begin{abstract}
Automotive radar sensors provide valuable information for advanced driving assistance systems (ADAS). Radars can reliably estimate the distance to an object and the relative velocity, regardless of weather and light conditions. However, radar sensors suffer from low resolution and huge intra-class variations in the shape of objects. Exploiting the time information (e.g, multiple frames) has been shown to help to capture better the dynamics of objects and, therefore, the variation in the shape of objects. Most temporal radar object detectors use 3D convolutions to learn spatial and temporal information. However, these methods are often non-causal and unsuitable for real-time applications. This work presents RECORD, a new recurrent CNN architecture for online radar object detection. We propose an end-to-end trainable architecture mixing convolutions and ConvLSTMs to learn spatio-temporal dependencies between successive frames. Our model is causal and requires only the past information encoded in the memory of the ConvLSTMs to detect objects. Our experiments show such a method's relevance for detecting objects in different radar representations (range-Doppler, range-angle) and outperform state-of-the-art models on the ROD2021 and CARRADA datasets while being less computationally expensive. 
\end{abstract}

\begin{IEEEkeywords}
Computer Vision and Pattern Recognition, Radar Object Detection,  Autonomous Driving,  Radar imaging 
\end{IEEEkeywords}

%%%%%%%%% BODY TEXT
\section{Introduction}
\label{sec:intro}
\IEEEPARstart{T}{oday}, most advanced driving assistance systems (ADAS) use cameras and LiDAR sensors to perceive and represent the surrounding environment. Cameras provide rich visual semantic information about the environment, while LiDARs provide high-resolution point clouds of surrounding targets. Large-scale image-based and LiDAR-based datasets \cite{coco_challenge, KITTI} have enabled the development of perception algorithms for object detection and segmentation. 
Despite their high resolution, camera and LiDAR sensors suffer from high sensitivity to harsh weather (fog, snow, rain) and bad light conditions (night or sunny).
On the contrary, radar operates at a millimetre wavelength, which can penetrate or diffract around tiny particles, making it a robust and crucial sensor for ADAS applications. Radar sensors also provide accurate localisation of surrounding targets and can estimate the velocity of vehicles and pedestrians in a single capture. However, the small number of public radar datasets and the lack of uniformity between them have slowed down research in deep learning for radar.

As shown in Figure \ref{fig:fmcw_radar}, radar data can be represented either as target lists or as raw data spectra (or tensors). Target lists are the default representation of radar data. They are obtained after several processing steps, including signal processing (Fourier transforms), threshold algorithms (CFAR \cite{blake1988cfar}), target tracking (Kalman filtering \cite{kalman}), and classification algorithms. Targets lists representation contains low-level information such as position $(x,y)$, azimuth $\theta$, relative speed $v_r$ and radar cross section $\sigma$ of targets. Classification or segmentation algorithms can be applied to these data, such as in \cite{palffy2022multi, 9190338, danzer20192d, liu_deep_2022}, to provide semantic information about the target.
However, the target lists have suffered several pre-processing steps. They do not contain all the initial information, which might lower the performance of classification or segmentation (ghost targets, sparse point clouds). Instead of radar target lists, it is possible to consider raw radar data tensors (range-doppler (RD), range-azimuth (RA) or range-azimuth-doppler (RAD) spectra) to exploit all the information available in the radar signal. In the last two years, several raw data datasets \cite{carrada, cruw_dataset, rebut2022raw} have been released to perform classification \cite{ulrich_deepreflecs_2021, patel_improving_2022}, object detection \cite{decourt_darod_2022, meyer2021graph, franceschi2022deep, rebut2022raw} or segmentation \cite{Ouaknine_2021_ICCV, kaul2020rss, rebut2022raw} on raw data tensors. However, most detection or segmentation methods are static. In other words, they use only a single image as input without exploiting the correlations among successive frames.

For automotive applications, time is key information which can be exploited to learn temporal patterns between successive frames in videos for example. In radar, given the modulation and the characteristic of the signal of FMCW (Frequency Modulated Continuous Wave) radars, the data includes temporal information (\textit{e.g.} Doppler effect), which is a crucial value in autonomous driving. The use of time in radar makes it possible to learn the dynamics of the objects held in the radar signal, handle the variation in the shape of the object over time, and reduce the noise between successive frames (induced by the movement of the surrounding object and the vehicle itself). Recent efforts have been made to exploit temporal relationships between raw data radar frames using multiple frames for detection or segmentation tasks. Mainly, Ouaknine \textit{et al.} \cite{Ouaknine_2021_ICCV} use 3D convolutions for multi-view radar semantic segmentation. \cite{wang2021rodnet} and \cite{ju2021danet} also take advantage of 3D convolutions for object detection on RA maps. In \cite{Major_2019_ICCV}, Major \textit{et al.} use ConvLSTM to detect cars in RA view and \cite{Li_2022_CVPR} processes sequences of two successive radar frames to learn the temporal relationship between objects.

This paper presents a new convolutional and recurrent neural network (CRNN) for radar spectra. Unlike most multi-frame radar object detectors, our model is causal, which means we only use past frames to detect objects. This characteristic is crucial for ADAS applications because such systems do not have access to future frames. To learn spatial and temporal dependencies, we introduce a model consisting of 2D convolutions and convolutional recurrent neural networks (ConvRNNs). Additionally, we use efficient convolutions and ConvRNNs (inverted residual blocks and Bottleneck LSTMs) to reduce the computational cost of our approach.
{O}ur model is end-to-end trainable and does not require pretraining or multiple training steps. We present a generic method that can process either RA, RD or RAD spectra and outperforms state-of-the-art architectures on different tasks. To our knowledge, this is the first fully convolutional recurrent network for radar spectra.

Section \ref{sec: radar_bkg} introduces fundamental radar signal processing. Then in Section \ref{sec: st-rod}, we describe our approach and the proposed model. \ref{sec: related} presents the prior art to this work. We present the results of our experiments on the ROD2021 dataset \cite{cruw_dataset} and the CARRADA \cite{carrada} dataset in Section \ref{sec:expe}. Finally, we discuss our results and conclude the paper in Section \ref{sec:discuss_concl}.
 
\section{Radar background}
\label{sec: radar_bkg}

\begin{figure}[!t]
    \centering
    \includegraphics[width=0.9\linewidth]{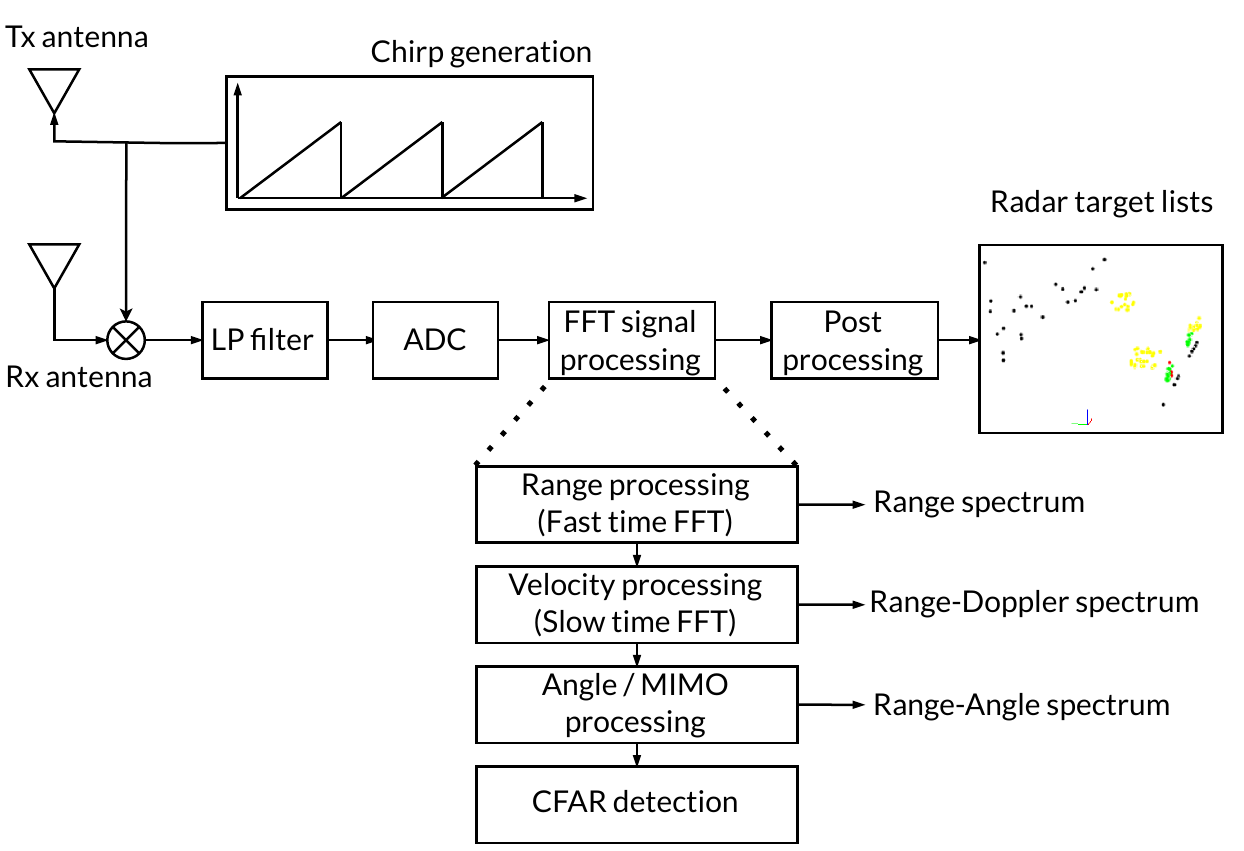}
    \caption{FMCW radar overview}
    \label{fig:fmcw_radar}
\end{figure}

\begin{figure*}[!t]
    \centering
    \includegraphics[width=1.0\linewidth]{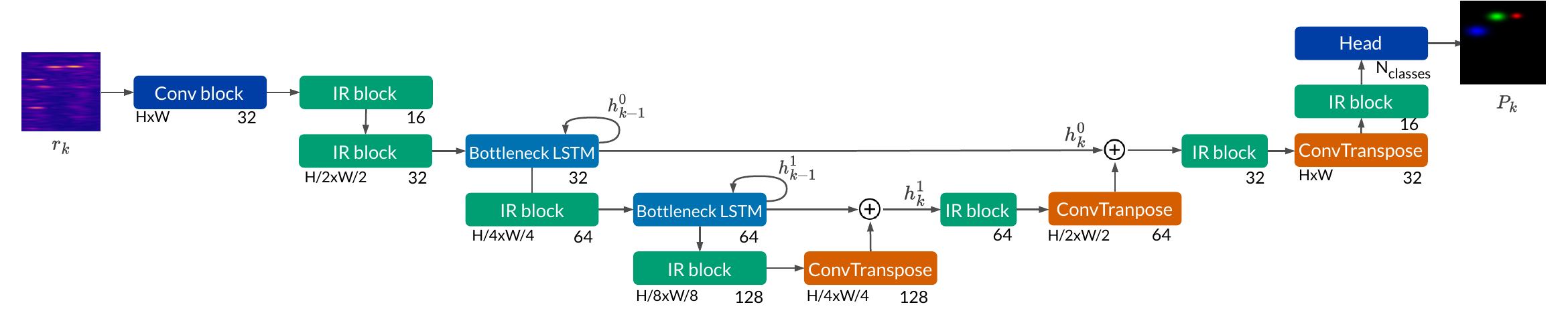}
    \caption{Model architecture (RECORD). Rounded arrows on \textit{Bottleneck LSTMs} stand for a recurrent layer. Plus sign stands for the concatenation operation. We report the output size (left) and the number of output channels (right) for each layer.}
    \label{fig:model_archi}
\end{figure*}

\noindent Radar is an active sensor that transmits radio frequency (RF) electromagnetic (EM) waves and uses the reflected waves from objects to estimate the distance, velocity, and angle of arrival of these targets \cite{radar_dsp}. Automotive radars emit a particular waveform called FMCW. An FMCW radar periodically transmits $P$ chirps (a frequency-modulated pulse whose frequency increases linearly with time) over $M_{Rx}$ receiving antennas and $M_{Tx}$ transmitting antennas to estimate the range, the velocity and the Direction-of-Arrival (DOA) of the targets. For the $m^{th}$ antenna, we express a single FMCW pulse as:
\begin{equation}
    s_n(t) = e^{j2\pi(f_c+0.5Kt)t} \: \: \: 0 \leq t\leq T,
    \label{eq:radar_fmcw}
\end{equation}
where $f_c$ is the carrier frequency, $K$ is a modulation constant, and $T$ is the duration of the chirp (fast time). 

As shown in Figure \ref{fig:fmcw_radar}, it is possible to estimate the distance to the target from the ADC data by applying a first discrete Fourier transform along the fast time index (\textit{i.e.} for every chirp). We obtain the velocity by computing a second discrete Fourier transform over the slow time index (\textit{i.e.} for each chirp index). This second Fourier transform allows measuring the frequency shift between received chirps resulting in the Doppler frequency and hence the velocity of targets. These two successive Fourier transforms result in a range-Doppler (or range-velocity) spectrum. 

From the RD spectrum, it is possible to obtain a list of targets which contains the position $x, y$, the radial velocity $v_r$, the DOA $\theta$ and the radar cross section $\sigma$. This is done by applying thresholding algorithms like CFAR \cite{blake1988cfar}, followed by the DOA estimation and post-processing steps (ego-motion compensation, Kalman filtering, classification) \cite{radar_dsp}. However, these operations reduce the amount of information in the signal.

In multiple receiving antenna scenarios, each antenna sees the reflected signal with a slight time delay. Computing a third discrete Fourier transform along the antenna array allows estimating targets' DOA and results in a range-azimuth-Doppler spectrum. Summing values along the Doppler dimension enable the computation of the range-angle spectrum. Compared to target lists, these representations contain much more information about the environment.

\section{Spatio-temporal radar object detector}
\label{sec: st-rod}

\noindent We aim to design a model to learn the implicit relationship between frames at different spatial and temporal levels recurrently. This section describes the architecture of our single-view spatio-temporal encoder and decoder. We also introduce a multi-view version of our model designed to learn spatial and temporal features in different views (\textit{e.g.} RD, AD, and RA) simultaneously.

\subsection{Problem formulation}
\label{subsec: formulation}

Let us consider a sequence $R$ of $N$ radar frames ranging from time $k-N+1$ to $k$ such as: $R = \{r_{k-N+1}, \ldots , r_{k}\}$. We aim to find the locations of every object in the scene at time $k$, $p_k$, based on the $N$ past frames. We define our model with two functions, the encoder $E$ and the decoder $D$. 

We consider a causal model, which means it uses only the past to predict the next time step. For each time step $k$ of the sequence, we consider a recurrent convolutional encoder taking as input the frame at time $k$ and a set of $I$ previous hidden states $H_{k-1} = \{h^0_{k-1}, \ldots, h^i_{k-1}, \ldots, h^{I-1}_{k-1}\}$  with $i$ the index of the recurrent unit if more than one are used. The encoder returns a set of features maps $F_k$ and a set of updated hidden states $H_{k} = \{h^0_{k}, \ldots, h^i_{k}, \ldots, h^{I-1}_{k}\}$ such that:
\begin{equation}
   E(r_k, H_{k-1}) = (F_k, H_k ).
\end{equation}
Because our encoder encodes the past $N$ frames recurrently to predict the position of objects at time step $k$, our decoder is a fully convolutional decoder that takes as input the encoder's updated hidden states $H_k$ (the memory) and the set of feature maps $F_k$ (spatio-temporal feature maps) such that: 
\begin{equation}
   D(F_k, H_k) = p_k.
\end{equation}

As we want to improve the classification accuracy more than the localisation accuracy, we use recurrent layers in the encoding phase only. In encoder-decoder architectures, the encoder learns to extract an abstract representation of the radar frame relative to the class while the decoder is used for localisation. Using recurrent layers only in the encoding phase allows the encoder to encode spatio-temporal relationships at the object level to improve the objects' representation.

\subsection{Spatio-temporal encoder}
\label{subsec: encoder}

An overview of our single-view architecture is shown in Figure \ref{fig:model_archi}. We propose a fully convolutional recurrent encoder (left part of Figure \ref{fig:model_archi}). In other words, our encoder mixes 2D convolutions and ConvRNNs. We use 2D convolutions to learn spatial information and reduce the size of inputs. To reduce the number of parameters of the model and its computation time, we use inverted residual (IR) bottleneck blocks from the MobileNetV2 \cite{Sandler_2018_CVPR} architecture instead of classic 2D convolutions for most of the convolutional layers of the model. IR bottleneck is a residual block based on depthwise separable convolutions that use an inverted structure for efficiency reasons. Then, we propose inserting ConvLSTM \cite{shi_convolutional_2015} cells between convolutional layers to learn the temporal relationship between frames. Similarly to the convolutions, we replace the classic ConvLSTM with an efficient one proposed in \cite{Liu_2018_CVPR} by Liu and Zhu called bottleneck LSTM. Contrary to a classic ConvLSTM, authors replace convolutions with depthwise-separable convolutions, which reduces the required computation by a factor of eight to nine. Additionally, $tanh$ activation functions are replaced by $ReLU$ activation functions. In this work, we use two bottleneck LSTMs, as a result, $I=2$. Such a layout enhances spatial features with temporal features and vice versa.

We follow the MobileNetV2 structure by first applying a full convolution to increase the number of channels followed by a single IR bottleneck block. Except for the first IR bottleneck block, we set the expansion rate $\gamma$ to four. Next, we apply two blocks composed of three IR bottleneck blocks followed by a bottleneck LSTM to learn spatio-temporal dependencies. Because the computational cost of bottleneck LSTMs is proportional to the input size, we use a stride of two in the first IR bottleneck block to reduce the input dimension. We insert these bottleneck LSTMs in the middle of the encoder to not alter the spatial information too much. Finally, we refine the spatio-temporal feature maps obtained from the bottleneck LSTMs by adding three additional IR bottleneck blocks. 

Because we treat data sequences, it is desirable to calculate normalisation statistics across all features and all elements for each instance independently instead of a batch of data (a batch can be composed of sequences from different scenes). As a result, we add layer normalisation before sigmoid activation on gates $o_t, i_t$ and $f_t$ in the bottleneck LSTM, and we adopt layer normalisation for all the layers in the model.

\subsection{Decoder}
\label{subsec: decoder}

As described in Section \ref{subsec: formulation}, our decoder is a 2D convolutional decoder which takes as input the last feature maps of the encoder (denoted $F_k$) and a set of two hidden states $H_k=\{h_k^0, h_k^1\}$. Our decoder is composed of three 2D transposed convolutions followed by a single IR block with an expansion factor $\gamma$ set to one, and a layer normalisation layer. Each transposed convolution block upsample the input feature map by two. Finally, we use two 2D convolutions as a classification/segmentation head (depending on the task) which projects the upsampled feature map onto the desired output. 

The U-Net architecture \cite{10.1007/978-3-319-24574-4_28} has popularised skip connections between the encoder and decoder. It allows precise localisation by combining high-resolution and low-resolution features. We, therefore, adopt skip connections between our encoder and our decoder to improve the localisation precision. {To prevent the loss of temporal information in the decoding stage, we use} the hidden states of each bottleneck LSTM (denoted by $h_k^0$ and $h_k^1$ in Figure \ref{fig:model_archi}) and concatenating them with the output of a transposed convolution operation to propagate in the decoder the temporal relationship learned by the encoder.

\begin{figure*}[!t]
    \centering
    \includegraphics[width=1.0\linewidth]{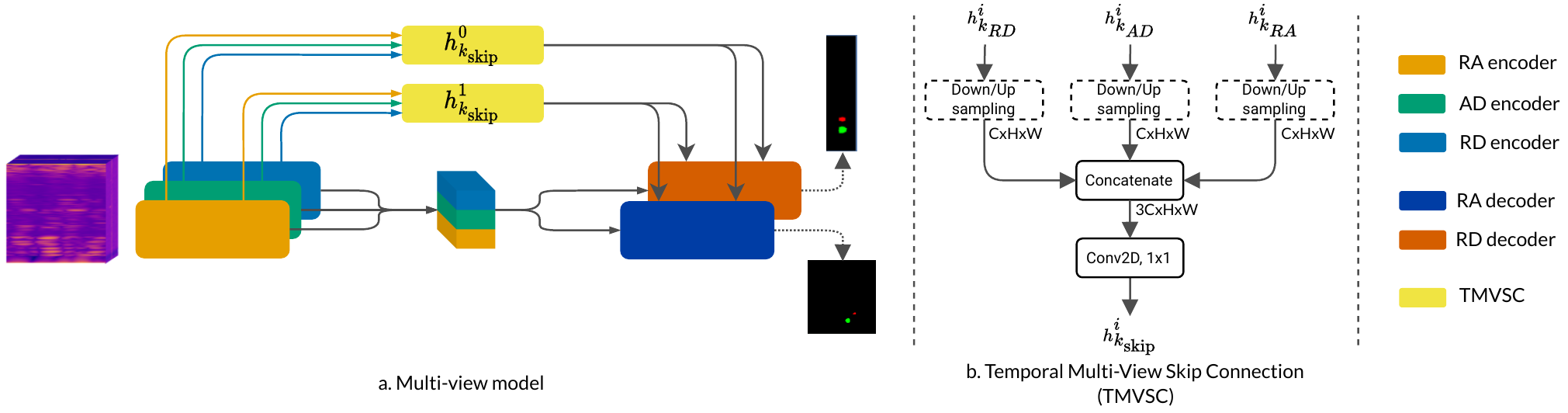}
    \caption{Multi-view model architecture (MV-RECORD). We use the encoder described in Figure \ref{fig:model_archi} for each view. Dashed boxes denote an optional operation applied only if the feature maps have different shapes. Gray arrows denote the same output.}
    \label{fig:mv_archi}
\end{figure*}

\subsection{Multi-view spatio-temporal object detector}
\label{subsec: mv-strod}

The preceding sections describe a spatio-temporal radar object detection architecture for single view inputs (\textit{i.e.} RA or RD). However, using more than one view to represent targets in their entirety might be desirable. In other words, to simultaneously find the position (distance, angle), the velocity and the class of targets. In this section, we propose to extend the previous architecture to a multi-view approach. We follow the paradigm of Ouakine \textit{et al.} \cite{Ouaknine_2021_ICCV} by replicating three times the encoder proposed in Section \ref{subsec: encoder} (one for RA view, one for RD view and one for AD view, see Figure \ref{fig:model_archi}). Then, the latent space of each view is concatenated to create a multi-view latent space. We use two decoders to predict objects' positions in all dimensions (RA and RD). One for the RA view and one for the RD view. The multi-view latent space is the input of these decoders.

In Section \ref{subsec: decoder}, we {use the hidden states of each bottleneck LSTMs for the skip connection to add} the temporal information in the decoding part. For the multi-view approach, we want to take advantage of the multi-view and the spatio-temporal approaches in the skip connections to supplement decoders with data from other views (\textit{e.g.} add velocity information in the RA view). Similarly to the multi-view latent space, we concatenate the hidden states from RD, RA and AD views. This concatenation results in a set of concatenated hidden states $H_k=\{h_{k_{skip}}^0, h_{k_{skip}}^1\}$. We describe the operation to obtain $H_k$ in Figure \ref{fig:mv_archi}b. We concatenate $H_k$ in the same way as in the single view approach. We call this operation Temporal Multi-View Skip Connections (TMVSC). Figure \ref{fig:mv_archi} illustrates the multi-view architecture we propose. 

\subsection{Training procedure}

\begin{figure}[!hb]
    \centering
    \subfloat[\label{subfig:many_to_one}]{
        \includegraphics[width=0.6\linewidth]{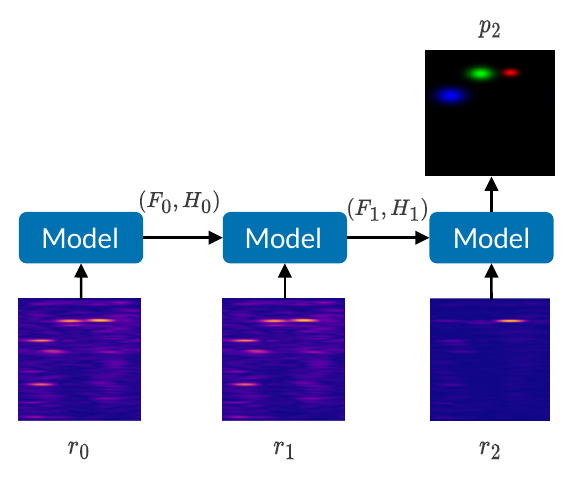}}
        \vfill
    \subfloat[\label{subfig:many_to_many}]{
        \includegraphics[width=0.6\linewidth]{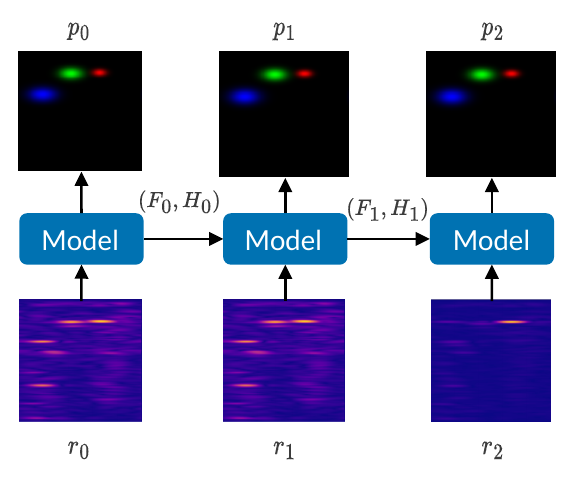}}
    \caption{Training procedures with $N=3$. (a) Buffer training procedure (many-to-one). (b) Online training procedure (many-to-many).}
    \label{fig:training_procedure}
\end{figure}

We propose two training methods to train RECORD and MV-RECORD: \textit{online} and \textit{buffer}, summarised in Figure \ref{fig:training_procedure}. Let us denote by $R = \{r_{k-N+1}, \ldots , r_{k}\}$ a sequence of $N$ radar frames ranging from time $k-N+1$ to $k$, $P = \{p_{k-N+1}, \ldots , p_{k}\}$ the objects' position in the sequence (the ground truth) and $\mathcal{L}$ the loss function we aim to minimise.

\paragraph{Buffer training}
We adopt a many-to-one paradigm when training using the \textit{buffer} approach. We train the model to predict only the position of the objects in the last frame $r_k$ as shown in Figure \ref{subfig:many_to_one}. Therefore, given a sequence of $N$ radar frames, we minimise the following loss function:
\begin{equation}
    \mathcal{L}(\hat{p}, p) = \mathcal{L}(\hat{p}_k, p_k) 
\end{equation}
where $k$ is the last time step of the sequence. \textit{Buffer} training forces the model to focus on a specific time window and to learn a global representation of the scene. However, in inference, the model must process $N$ frames sequentially to make a prediction. Therefore, we propose to train the model differently using a many-to-many paradigm to improve the model's efficiency in inference.

\paragraph{Online training}
We adopt a many-to-many paradigm when training using the \textit{online} approach. We train the model to predict the position of the objects for every frame in the sequence $R$ as shown in Figure \ref{subfig:many_to_many}. Therefore, given a sequence of $N$ radar frames, we minimise the following loss function:
\begin{equation}
    \mathcal{L}(\hat{p}, p) = \sum_{k=1}^N\mathcal{L}(\hat{p}_k, p_k) 
\end{equation}
\textit{Online} training pushes the model to use previous objects' positions to make a new prediction. It encourages the model to keep only relevant information from the previous frames. \textit{Online} training requires training with longer sequences but allows data processing one by one (no buffer) in inference. In contrast to the \textit{buffer} approach, the hidden states are not reset in inference.

\section{Related work}
\label{sec: related}

\subsection{Sequential object detection and segmentation in computer vision}
 
Object detection and segmentation are fundamental problems in computer vision. However, the majority of object detection and segmentation algorithms have been developed on static images. For some applications (robotics, autonomous driving, earth observation), processing sequences of images is desirable. Due to motion blur or object occlusion, it is sub-optimal to directly apply classical object detectors, or segmentation algorithms \cite{deeplabv3plus2018} on successive frames. To exploit the temporal information in sequences, optical flows \cite{zhu2017flow,yu2022batman}, recurrent networks with or without convolutions \cite{Liu_2018_CVPR, li2018recurrentretina, pfeuffer2019semanticconvlstm, ventura2019rvos, zhang2019modelingonline}, attention \cite{garnot2021panoptic, yu2022batman}, transformers \cite{yu2022batman, duke2021sstvos}, aggregation methods \cite{chen2020memorymega} or convolutions \cite{xiao2018videostmemory, bertasius2018objectstsampling} were widely used. 

In \cite{li2018recurrentretina} and \cite{Liu_2018_CVPR}, authors propose to transform the SSD object detector in a recurrent model by adding ConvLSTM cells on the top of the regression and classification head for \cite{li2018recurrentretina} and between the last feature map of the feature extractor and the detection head for \cite{Liu_2018_CVPR}. However, these models predict bounding boxes (not segmentation masks) and learn to find temporal relationships only between feature maps. In \cite{ventura2019rvos}, authors present a recurrent model for one-shot and zero-shot video object instance segmentation. Contrary to previous methods and ours, they use a fully recurrent decoder composed of upsampling ConvLSTM layers to predict instance segmentation masks. Another approach proposed by Sainte Fare Garnot  and Landrieu in \cite{garnot2021panoptic} consists in using temporal self-attention to extract multi-scale spatio-temporal features for panoptic segmentation of satellite image time series.

Despite some models being trained online (\textit{i.e.} no access to future frames during training), \cite{Liu_2018_CVPR, pfeuffer2019semanticconvlstm,li2018recurrentretina}, some models using a sequence of images are non-causal in inference and use the video in its entirety. Thus, \cite{zhang2019modelingonline} propose a causal recurrent flow-based method for online video object detection. Their method only uses the past and the current frame from a memory buffer and learns short-term temporal context using optical flow and long-term temporal context using a ConvLSTM. Nevertheless, this method needs to learn optical flow to get accurate results, which is not possible in radar.

\subsection{Sequential object detection and segmentation in radar}

In radar, object detection or segmentation algorithms \cite{schumann2018semantic, gao2020ramp, franceschi2022deep, zhang2021raddet, decourt_darod_2022} that do not use time suffer from low performances for similar classes such as pedestrians and bicyclists \cite{Ouaknine_2021_ICCV, decourt_darod_2022}. According to the Doppler principle, motion information is held in the radar signal and should help to differentiate a pedestrian (non-rigid body, motion information widely distributed), and a car (rigid body, more consistent motion information) \cite{wang2021rodnet}. 3D convolutions are primarily used in radar to learn spatio-temporal dependencies between frames. Methods such as \cite{wang2021rodnet, gao2020ramp, ju2021danet, zheng2021scene} adopt 3D encoder-decoder architectures where they predict the position of objects for $N$ successive frames. Despite their performances, these methods require a buffer of $N$ frames in memory to work and are not really online methods in inference, as the convolutional kernel is applied over past and future frames. On the contrary, our approach doesn't access future frames either in training or inference. Additionally, the number of parameters of models using 3D convolutions is huge for real-time applications (34.5M for RODNet-CDC \cite{wang2021rodnet}, 104M for RAMP-CNN \cite{gao2020ramp}). Consequently, Ju \textit{et al.} \cite{ju2021danet} introduced Dimension Apart Module (DAM), a lightweight module for spatio-temporal features extraction on RA maps that can be integrated into U-Net style network architecture. Alternatively, Ouaknine \textit{et al.} propose in \cite{Ouaknine_2021_ICCV} to use 3D convolutions to encode the spatial information of the $N$ past frames in an online setting. Similarly to other 3D convolutions-based methods, TMVA-Net \cite{Ouaknine_2021_ICCV} has a lot of parameters compared to our approach. 

Kaul \textit{et al.} \cite{kaul2020rss} propose a model without 3D convolutions where the time information is stacked in the channel dimension. \cite{niederlohner2022self} aggregates point clouds of different time steps to increase the resolution of the radar point cloud. More recently, Liu \textit{et al.} \cite{liu_deep_2022} propose an approach inspired by the transformer architecture and based on computer vision-based feature extractors to exploit temporal dependencies between objects in two successive frames. Finally, Major \textit{et al.} \cite{Major_2019_ICCV} follow \cite{broad2018recurrent} by using a ConvLSTM over the features of a multi-view convolutional encoder. Even though this model is similar to ours, the LSTM cell is applied only to the learned cartesian output before the detection head, and the proposed model only detects cars. Additionally, this model produces bounding boxes, which is not very accurate for radar, is not end-to-end trainable and requires pre-training of a non-recurrent version of it before.

\section{Experiments}
\label{sec:expe}

\subsection{Single view object detection}
\label{subsec: svresults}

\begin{table*}[!t]
\begin{center}
\caption{Results obtained on the test set of the ROD2021 challenge for different driving scenarios (PL: Parking Lot, CR: Campus Road, CS: City Street and HW: Highway). We report {the best results} over five different seeds. The best results are in bold, and the second bests are underlined.}
\label{tab:cruw_res}
\resizebox{\textwidth}{!}{%
\begin{tabular}{@{}lllllllllllccc@{}} \toprule
\multicolumn{1}{c}{\multirow{2}{*}{Model}} & \multicolumn{5}{c}{AP} & \multicolumn{5}{c}{AR} & \multirow{2}{*}{Params (M)} & \multirow{2}{*}{GMACS} & \multirow{2}{*}{Runtime (ms)} \\ \cmidrule(lr){2-6} \cmidrule(lr){7-11}
\multicolumn{1}{c}{} & \multicolumn{1}{c}{Mean} & \multicolumn{1}{c}{PL} & \multicolumn{1}{c}{CR} & \multicolumn{1}{c}{CS} & \multicolumn{1}{c}{HW} & \multicolumn{1}{c}{Mean} & \multicolumn{1}{c}{PL} & \multicolumn{1}{c}{CR} & \multicolumn{1}{c}{CS} & \multicolumn{1}{c}{HW} & \multicolumn{2}{l}{} \\ \midrule
RECORD (buffer, ours) & {\ul 72.8} & 95.0  &  \underline{67.7} & 48.3 & \textbf{77.4} & \textbf{82.8} & \underline{96.7} & 73.9 & \textbf{72.8} &\textbf{81.7} & 0.69 & 5.0 & 61.6\\
RECORD (online, ours) & \textbf{73.5} & \textbf{96.4} & \textbf{72.5} & 49.9 & {\ul 72.5} & {\ul 81.2} &  96.4 & {\ul 78.1} & 68.8 & {\ul 77.6} & 0.69 & 0.95 & 6.2\\
RECORD (no lstm, multi)  & 65.5 & 89.9 & 57.3 & 43.1 & 68.9 & 78.9 & 93.1 & 68.2 & \underline{71.5} & 75.7 & 0.47 & 0.76 & 5.8\\
RECORD (no lstm, single)  & 59.5 & 85.7 & 48.5 & 39.11 & 64.4 & 75.1 & 90.8 & 62.4 & 68.9 & 69.6 & 0.44 & 0.35 & 5.7\\
DANet \cite{ju2021danet} & 71.9 & 94.7 & 65.7 & \textbf{51.9} & 70.0 & 80.7 & 96.2 & 75.1 & \textbf{72.8} & 73.0 & 0.74 & 9.1 & 21.1 \\
%DANet-C \cite{ju2021danet} & 69.3 & 95.5 & 63.7 & 46.5 & 66.8 & 78.4 & 96.6 & 71.7 & 67.9 & 70.5 & 0.74 & 9.1 & - \\
UTAE \cite{garnot2021panoptic} & 68.4 & 92.1 & 67.4 & \underline{51.4} & 65.5 & 78.4 & 94.6 & 74.0 & 69.7 & 70.0 & 0.79 & 4.1 & 7.5 \\
\textcolor{black}{T-RODNet \cite{trodnet}} & \textcolor{black}{69.9} & \textcolor{black}{\underline{95.6}} & \textcolor{black}{\textbf{72.5}} & \textcolor{black}{48.2} & \textcolor{black}{63.7} & \textcolor{black}{79.5} & \textcolor{black}{\textbf{97.2}} & \textcolor{black}{\textbf{79.1}} & \textcolor{black}{70.2} & \textcolor{black}{67.2} & \textcolor{black}{159.7} & \textcolor{black}{91.7} & \textcolor{black}{74.0} \\
\bottomrule
\end{tabular}%
}
\end{center}
\end{table*}

\paragraph{Dataset} We prototype and train RECORD on the ROD2021 challenge dataset\footnote{https://www.cruwdataset.org/rod2021}, a subset of the CRUW dataset \cite{cruw_dataset}.  Due to its high frame rate (30 fps), this dataset is well-suited to evaluate the temporal models. The ROD2021 dataset contains 50 sequences (40 for training and 10 for testing) of synchronised cameras and raw radar frames. Each sequence contains around 800-1700 frames in four different driving scenarios, \textit{i.e.}, parking lot (PL), campus road (CR), city street (CS), and highway (HW). 

The provided data of the ROD2021 challenge dataset are pre-processed sequences of RA spectra (or maps). Annotations are confidence maps (ConfMaps) in range-azimuth coordinates that represent object locations (see Figure \ref{fig:model_archi}). According to \cite{wang2021rodnet} one set of ConfMaps has multiple channels, each representing one specific class label, \textit{i.e.}, car, pedestrian, and cyclist. The pixel value in the $cls$-th channel represents the probability of an object with class $cls$ occurring at that range-azimuth location. We refer the reader to \cite{wang2021rodnet} for more information about ConfMaps generation and post-processing. RA spectra and ConfMaps have dimensions $\text{128}\times\text{128}$.

\paragraph{Evaluation metrics} We use the metric proposed in \cite{wang2021rodnet} to evaluate the models on the ROD2021 challenge datasets. In image-based object detection, intersection over union (IoU) is mostly used to estimate how close the prediction and the ground truth (GT) are. For our single-view approach, as we predict the location of objects, we utilise the object location similarity (OLS) to match detection and GT. The OLS is defined as:
\begin{equation}
    OLS = \exp{\frac{-d^2}{2(s\kappa_{cls})^2}}
\end{equation}
where $d$ is the distance (in meters) between the two points in the RA spectrum, $s$ is the object distance from the radar sensor (representing object scale information) and $\kappa_{cls}$ is a per-class constant that describes the error tolerance for class $cls$ (average object size of the corresponding class). First, OLS is computed between GT and detection. Then the average precision (AP) and the average recall (AR) are calculated using different OLS thresholds ranging from 0.5 to 0.9 with a step of 0.05, representing different localisation error tolerance for the detection results. In the rest of this section, AP and AR denote the average precision and recall for all the thresholds. 

 \paragraph{Evaluation procedure}
 In the ROD2021 dataset, annotations of test sequences are unavailable.To train and evaluate our models and the baselines similarly, we selected 36 sequences out of 40 to train the model and four for validation. We use the validation set for early stopping. Once the models are trained, we test them on the test set. Finally, we update the prediction on the ROD2021 evaluation platform to evaluate the performance of each model. As for the ROD2021 challenge, the evaluation is done for 70\% of the test set.

\paragraph{Competing methods} We compare our approach with several radar-based and image-based methods using sequences of multiple radar frames. For the radar-based approach, we first benchmark our model against DANet\footnote{{The original implementation is not available so we implement it according to author's guidelines. We do not use test augmentation and ensemble learning in this paper.}} \cite{ju2021danet}, a 3D convolutional model which won the ROD2021 challenge. Because image-based models are too heavy for our application, we finally contrast our recurrent approach against the attention-based model UTAE \cite{garnot2021panoptic}, which is lighter than image-based approaches and which we can use causally. We found that decreasing the number of channels of UTAE and changing the positional encoding improved the performances (see Table~\ref{tab:utae}). We also consider {two variants of our model without LSTMs, one using the time along the channel dimension (\textit{no lstm, multi)} and one using a single frame (\textit{no lstm, single)}.}

\paragraph{Experimental settings}
We use sequences of 32 frames for the \textit{online} training. For validation and testing, frames are processed one by one. For the \textit{buffer} training, we use sequences of 12 frames in both training and evaluation. Here we reset the hidden states every 12 frames. We use this buffer approach for fair comparison with other baselines that also use a buffer, although it is less efficient than the online approach.  We optimise our model using the Adam optimiser with learning rate ($1\times10^{-3}$) for the buffer method {and $3\times10^{-4}$ for the online method}. We decay the learning rate exponentially by a factor of 0.9 every ten epochs. We train all the models using a binary cross-entropy loss.

We use an early stopping strategy to stop training if the model does not improve for seven epochs. To avoid overfitting, we use a stride of four {for the buffer model and eight for the online model} (i.e., how many frames the model skips between each training iteration) in the training dataset. The stride is set to one in validation and testing as we process data on the fly. We apply different data augmentation techniques during training, such as horizontal, vertical and temporal flipping. We use these settings for all the baselines, except for DANet where we use the settings recommended by the authors. All the models were implemented using the Pytorch Lightning\footnote{https://www.pytorchlightning.ai/} framework and trained on an NVIDIA Quadro RTX 8000 GPU. We run all the models with five different seeds and report the {best} results in the next paragraph. 

\paragraph{Results} Table~\ref{tab:cruw_res} presents the results of our model and the baselines on the test set of the ROD2021 dataset. Our recurrent approaches generally outperform baselines for both AP and AR metrics; this remains true for most scenarios. {The online version of RECORD obtains} the best trade-off between performances and computational complexity (parameters, number of multiplications and additions and runtime). {Despite having less GMACs than UTAE and DANet, the buffer version of RECORD is the slowest one among all the models. Indeed, for each new frame we need to process the 11 previous ones, which is inefficient. Results show that the online version should be preferred for real-time applications.} Additionally, RECORD methods exceed 3D and attention-based methods on static scenarios such as parking lot (PL) and campus road (CR). In PL and CR scenarios, the radar is static and the velocity of targets varies a lot, our recurrent models seem to learn variations of the target's speed better than other approaches. Surprisingly the attention-based method UTAE, initially designed for the segmentation of satellite images, obtains very competitive results with our method and the DANet model. We notice that the approach using the time in the channel dimension reaches lower AP and AR than their counterpart, which explicitly uses time as a new dimension. Finally, training our model without the time and using only a 2D backbone (\textit{no lstm, single)} obtain the lower performance on the test set. 
\begin{table}[!t]
\begin{center}
\caption{Comparison of different types of ConvRNN. We train all the models with the same loss and hyperparameters. Bottleneck LSTM achieves the best AP while having fewer parameters and GMACS.}
\label{tab:convrnn}
\resizebox{\columnwidth}{!}{%
\begin{tabular}{@{}cllcc@{}} \toprule
ConvRNN type & AP                      & AR                      & Params (M) & GMACS \\ \midrule
Bottleneck LSTM \cite{Liu_2018_CVPR} & \textbf{69.8 $\pm$ \small 2.2} & {\ul 80.2 $\pm$ \small 1.5}    & 0.69 & 5.0 \\
ConvLSTM  \cite{shi_convolutional_2015}   & 66.63 $\pm$ \small 3.28 & 79.36 $\pm$ \small 2.39 & 1.0 & 11.6          \\
ConvGRU  \cite{ConGRU}       & {\ul 69.7 $\pm$ \small 2.4}    & \textbf{81.2 $\pm$ \small 1.4} & 0.94 & 9.8 \\ \bottomrule
\end{tabular}%
}
\end{center}
\end{table}

\begin{table}[!t]
\begin{center}
\caption{Comparison of different types of skip connections. Results are averaged over 5 different seeds on the ROD2021 test set. Concatenation is the RECORD model, addition stand for a model where we add the output of transposed convolutions to $h_k^i$, and no skip connection stands for a model without skip connections. }
\label{tab:skip}
\resizebox{\columnwidth}{!}{%
\begin{tabular}{@{}cllc@{}} \toprule
Skip connection     & AP                      & AR                      & Params (M)\\ \midrule
Concatenation       & \textbf{69.8 $\pm$ \small 2.2} & {\ul 80.2 $\pm$ \small 1.5} & 0.69         \\
Addition            & {\ul 64.4 $\pm$ \small 5.3} & \textbf{80.5 $\pm$ \small 1.1} & 0.58         \\
No skip connections & 63.7 $\pm$ \small 6.2 & 78.7 $\pm$ \small 3.5 &  0.58 \\ \bottomrule
\end{tabular}%
}
\end{center}
\end{table}

\paragraph{Ablation studies}
We demonstrate the relevance of using bottleneck LSTMs instead of classic ConvGRUs or ConvLSTMs in Table~\ref{tab:convrnn}. Bottleneck LSTMs reduce the number of parameters and GMACS and achieve higher AP and AR than classic ConvRNNs. Additionally, we show in Table~\ref{tab:skip} the AP and the AR of our model with different skip connections. We show that concatenating temporal features with spatial features of the decoder (i.e., our RECORD model) reaches better AP and AR than a method without skip connections, or one where we add the temporal features to the spatial features of the decoder. Nevertheless, the concatenation of features increases the number of parameters and the number of GMACS of the model compared to other approaches.

\paragraph{UTAE performances improvement} 
Table~\ref{tab:utae} {depicts the performance improvement of UTAE \cite{garnot2021panoptic} model with and without positional encoding and with a modified number of channels in the encoder and the decoder. We modify the number of channels of UTAE to match our architecture. We found that decreasing the model size and using positional encoding improve the model's performance. We define positional encoding as the time between the first and the $k^{th}$ frames.}

\begin{table}[!t]
\begin{center}
\caption{Performances improvement of UTAE model with and without positional encoding and with the default architecture (underlined line). Results are obtained on the test set and on a single seed. }
\label{tab:utae}
\resizebox{\columnwidth}{!}{%
\begin{tabular}{cccccc} \toprule
\multicolumn{2}{c}{\# channels} & \multicolumn{1}{l}{\multirow{2}{*}{Pos. enc.}} & \multicolumn{1}{l}{\multirow{2}{*}{AP}} & \multicolumn{1}{l}{\multirow{2}{*}{AR}} & \multicolumn{1}{l}{\multirow{2}{*}{Params (M)}} \\ \cmidrule(lr){1-2}
Encoder & Decoder & \multicolumn{1}{l}{} & \multicolumn{1}{l}{} & \multicolumn{1}{l}{} & \multicolumn{1}{l}{} \\ \midrule
16, 32, 64, 128 & 16, 32, 64, 128 & Yes & \textbf{68.4} & \textbf{78.4} & 0.79 \\
16, 32, 64, 128 & 16, 32, 64, 128 & No & 46.9 & 64.3 & 0.79 \\
{\ul 64, 64, 64, 128} & {\ul 32, 32, 64, 128} & Yes & 60.8 & 77.9 & 1.1 \\ \bottomrule
\end{tabular}%
}
\end{center}
\end{table}

\subsection{Multi-view semantic segmentation}
\label{subsec: mvresults}

\begin{table*}[!t]
\begin{center}
\caption{Results on the multi-view approach on CARRADA dataset. MV-RECORD stands for our multi-view approach. RECORD* stands for a single-view approach. The best results are in bold, and the second bests are underlined.}
\label{tab:carrada_res}
%\resizebox{\textwidth}{!}{%
\begin{tabular}{@{}cccllllccc@{}} \toprule
\multicolumn{2}{c}{\multirow{2}{*}{Model}} & \multicolumn{5}{c}{IoU} & \multirow{2}{*}{Params (M)} & \multirow{2}{*}{GMACS} & \multirow{2}{*}{Runtime (ms)}\\ \cmidrule(lr){3-7}
\multicolumn{2}{c}{} & mIoU & Bg & Ped & Cycl & Car & \\ \midrule
\multirow{5}{*}{RA} 
& MV-RECORD (buffer, ours) & \textbf{44.5} & 99.8 & {\ul 24.2} & \textbf{20.1} & {\ul 34.1} & 1.9 & 22.2 & 281.8  \\
& MV-RECORD (online, ours) & {\ul 42.4} & 99.8 & 22.1 & {\ul 11.1} & \textbf{36.4} & 1.9 & 3.7  & 56.6 \\
 & RECORD* (buffer, ours) & 34.8 & 99.7 & 10.3 & 1.4 & 27.7 & 0.69 & 8.2 & 116.1\\
 & RECORD* (online, ours) & 36.3 & 99.8 & 12.1 & 3.1 & 30.4 & 0.69 & 2.38  & 29.6 \\
 & TMVA-Net \cite{Ouaknine_2021_ICCV} & 41.3 &  99.8 & \textbf{26.0} & 8.6 & 30.7 & 5.6 & 98.0 & 21.8\\
 & MV-Net \cite{Ouaknine_2021_ICCV} & 26.8 & 99.8 & 0.1 & 1.1 & 6.2 & 2.4 & 53.3 & 18.8 \\ \midrule[0.05pt]
\multirow{5}{*}{RD}
& MV-RECORD (buffer, ours) & \textbf{63.2} & 99.6 & \textbf{54.9} & \textbf{39.3} & {\ul 58.9} &  1.9 & 22.2 & 281.8\\
& MV-RECORD (online, ours) & 58.5 & 99.7 & 49.4 & 26.3 & 58.6 & 1.9 & 3.7  & 56.6 \\
& RECORD* (buffer, ours) & 58.1 & 99.6 & 46.6 & 28.6 & 57.5 & 0.69 & 6.1 & 58.5  \\
& RECORD* (online, ours) & {\ul 61.7} & 99.7 & 52.1 & {\ul 33.6} & \textbf{61.4} & 0.69 & 0.59 & 13.3 \\
 & TMVA-Net \cite{Ouaknine_2021_ICCV} & 58.7 & 99.7 & {\ul 52.6} & 29.0 & 53.4 & 5.6 & 98.0 & 21.8 \\
 & MV-Net \cite{Ouaknine_2021_ICCV} & 29.0 & 98.0 & 0.0 & 3.8 & 14.1 & 2.4 & 53.3 & 18.8\\ \bottomrule
\end{tabular}%
%}
\end{center}
\end{table*}

\paragraph{Dataset} To demonstrate the relevance of our method, we train our model for multi-view object segmentation on the CARRADA dataset \cite{carrada}. The CARRADA dataset contains 30 sequences of synchronised cameras and raw radar frames recorded in various scenarios with one or two moving objects. The CARRADA dataset provides RAD tensors and semantic segmentation masks for both RD and RA views. Contrary to the CRUW dataset, the CARRADA dataset only contains simple driving scenarios (static radar on an airport runway). The frame rate is 10Hz. The objects are separated into four categories: pedestrian, cyclist, car and background. The RAD tensors have dimensions $\text{256}\times\text{256}\times\text{64}$ and the semantic segmentation masks have respectively dimensions $\text{256}\times\text{256}$ and $\text{256}\times\text{64}$ for the RA and the RD spectra. For training, validation and testing, we use the dataset splits provided by the authors. 

\paragraph{Evaluation metrics} {W}e evaluate our multi-view model using the intersection over union (IoU). IoU is a common evaluation metric for semantic image segmentation, which quantifies the overlap between the target mask T and the predicted segmentation mask P. For a single class, IoU is defined as:
\begin{equation}
    IoU = \big| \frac{T \cap P}{T \cup P} \big|.
\end{equation}
We then average this metric over all classes to compute the mean IoU (mIoU). 

%Another metric used in semantic segmentation is the Dice score. Dice score is defined as:
%\begin{equation}
%    Dice = \frac{2 |T \cap P|}{|T| \cup |P|}.
%\end{equation}
%Dice score is similar to F1-score, which can be seen as the harmonic mean of precision and recall. The Dice score represents overall performance, while the IoU evaluates worst-case performance. We average the Dice score over all classes (mDice) to assess the global performance of the model.

\paragraph{Competing methods} We compare our multi-view model with state-of-the-art multi-view radar semantic segmentation models, namely MV-Net and TMVA-Net \cite{Ouaknine_2021_ICCV}. Additionally, we train a buffer and an online single-view variant of our RECORD model. We train two different models, one for the RA view and one for the RD view.

\paragraph{Experimental settings} {As for the ROD2021 dataset, we use two evaluation settings for MV-RECORD: online and buffer.} The CARRADA dataset has a significantly lower frame rate than the ROD2021 dataset. In order to match the same time as a single view model, we set the number of input frames to five {for the buffer variant and ten for the online one,} which corresponds to a time of respectively 0.5  and 1 second. We set the batch size to eight and optimise the model using Adam optimiser with a learning rate of $1 \times 10^{-3}$ for both buffer and online methods except for the online multi-view model where the learning rate is set to $3\times 10^-4$.

We decay exponentially the learning rate every 20 epochs with a factor of 0.9. We use a combination of a weighted cross-entropy loss and a dice loss with the recommended parameters described in \cite{Ouaknine_2021_ICCV} to train our model as we find it provides the best results. To avoid overfitting, we apply horizontal and vertical flipping data augmentation. We also use an early stopping strategy to stop training if the model's performance does not improve for 15 epochs. Training multi-view models is computationally expensive (around six days for TMVA-Net and five days for ours). As a result, we train models using the same seed as the baseline for a fair comparison. We use the pre-trained weights of TMVA-Net and MV-Net to evaluate baselines. 

\paragraph{Results} Table~\ref{tab:carrada_res} shows the results we obtain on the CARRADA dataset. Our multi-view approaches beat the state-of-the-art model TMVA-Net on the multi-view radar semantic segmentation task while using two times fewer parameters and requiring significantly fewer GMACS. Our approach seems to correctly learn the variety of objects' shapes without complex operations such as the atrous spatial pyramid pooling (ASPP) used in TMVA-Net. We notice that using recurrent units instead of 3D convolutions in a multi-view approach significantly helps to improve the classification of bicyclists and cars, especially on the RA view, where we double the IoU for bicyclists compared to TMVA-Net. However, bicyclists and pedestrians are very similar classes, and improving the detection performance of bicyclists leads to a loss in the detection performance of pedestrians for the RA view. In the RD view, MV-RECORD models outperform the TMVA-Net approach for all classes. We notice a huge gap in RA view performances between the CARRADA dataset and the CRUW dataset, as well as between the two views of the CARRADA dataset. We hypothesise that the small frame rate of the CARRADA dataset might cause these differences. Indeed, the RD view contains Doppler information, enabling one to learn the dynamics of targets. However, the RA view might not contain as much motion information as in the ROD2021 dataset, where the frame rate is higher, allowing the network to learn the dynamics of targets even in the RA view. 
{Unfortunately, we cannot share the same analysis for the online multi-view approach. Compared to the results on the ROD2021 dataset, where the online approach performs better than the buffer one, we could not find proper training settings for the online multi-view model. Despite MV-RECORD online reaching higher IoU than TMVA-Net on the RA view, this model performs similarly with TMVA-Net on the RD view but has significantly lower IoU than the MV-RECORD buffer approach. We think these differences are mostly optimisation problems. Indeed, we show the online training outperforms the buffer training when using a single view on both the ROD2021 (Table~\ref{tab:cruw_res}) and the CARRADA dataset. Especially on the RD view, our single view and online model outperforms TMVA-Net without using the angle information, with 8 times fewer parameters and less computations. This confirms that the low frame rate of the CARRADA dataset limits the motion information that the recurrent layers can learn.} {Finally, despite having fewer GMACS and parameters than TMVA-Net, our multi-view model ({buffer}) is much slower in inference than TMVA-Net and is unsuitable for real-time application. The {online} version is faster and should be preferred for real-time applications. Decreasing the size of the feature maps in the early layer of the network might help to increase the inference speed of the model. Also, we notice using a profiler that the LayerNorm operation takes up to 90\% of the inference time for the multi-view models and up to 70\% of the inference time for the single-view models. Replacing layer normalisation with batch normalisation should speed up the runtime of our approaches. Given the good results of the single-view approach (especially for the RD view), we recommend using our model for single-view inputs, as RECORD was originally designed for this single-view object detection.}

\subsection{Discussion}

\paragraph{The difference with the results in the DANet paper \cite{ju2021danet}}
Experiments in Section \ref{subsec: svresults} show that DANet produces a 71.9 AP and 79.5 AR which is different from the results announced in the original paper. The code of DANet being unavailable, we implemented it according to the author's guidelines. Although we obtained the same number of parameters announced for the DAM blocks, our implementation has 740k parameters instead of the 460k announced in the paper. Beyond the implementation, the training and evaluation procedure in our paper is different from the one in the DANet \cite{ju2021danet} paper. While DANet is trained on the entire training set, we trained it on 36 carefully chosen sequences for a fair comparison with other models. Also, DANet authors use the following techniques when testing the model to improve the performance: test-time augmentation (TTA), ensemble models and frame averaging. Because DANet predicts frames by a batch of 16 with a stride of four, the authors average the overlapping frames (12 in total) in inference. Together, those techniques boost the performance of DANet around ten points, according to the ablation studies in DANet's paper, which is coherent with the gap between our scores and the ones from DANet paper. While applying TTA, ensemble models and training all the models using all the sequences would certainly also improve the global performance of all the models in Table \ref{tab:cruw_res}, we preferred comparing the architectures on a simpler but fair evaluation.

\paragraph{General discussion}
Here we discuss and analyse the results presented in Sections \ref{subsec: svresults} and \ref{subsec: mvresults}. First, radar data differs from LiDAR and images. The most critical differences being 1) the data is simpler in terms of variety, size, and complexity of the patterns; and 2) the datasets {are} smaller. We thus believe that our lighter architectures are flexible enough, while being less sensitive than huge backbones and less prone to overfitting for radar data. This mainly explains why Bottleneck LSTMs perform better than ConvLSTMs/ConvGRUs (see Table~\ref{tab:convrnn}). Also, we think convolutional LSTMs are more adapted to radar sequences because 1) convLSTMs learn {long-term} spatio-temporal dependencies at multiple scales, which 3D convolution cannot do because of the limited size of the temporal kernel; 2) LSTMs can learn to weigh contributions of different frames which can be seen as an adaptive frame rate depending on the scenarios and the speed of vehicles; 3) Hidden states keep the position/velocity of objects in previous frames in memory and use it to predict the position in the next time steps. Indeed, we show that, except for MV-RECORD, which is hard to optimise, online methods generally perform better than buffer ones while having lower computational cost (GMACs and inference time).

To conclude, although multi-view methods are interesting for research purposes, we find them difficult and long to optimise. Current radar generations generally detect targets in the range-Doppler view and compute the direction of arrival for each detected target to save computation time. As the RAD cube is cumbersome to compute and store in memory, we suggest using our model on single-view inputs (RD or RA), depending on the desired application.

\section{Conclusion}
\label{sec:discuss_concl}

\noindent In this work, we tackled the problem of online object detection for radar using recurrent neural networks. Contrary to well-known radar object detectors, which use a single frame to detect objects in different radar representations, we learn spatial and temporal relationships between frames by leveraging characteristics of FMCW radar signal. We propose a new architecture type that iteratively learns spatial and temporal features throughout a recurrent convolutional encoder. We designed an end-to-end, efficient, causal and generic framework that can process different types of radar data and perform various detection tasks (key point detection, semantic segmentation). Our methods outperform competing methods on both CARRADA and ROD2021 datasets. Notably, our models help distinguish pedestrians and cyclists better and learn the target variations better than 3D approaches. 

The main challenge in the near future will be to embed this model onboard real cars. This will require a more involved training, a quantisation of the model, and improved data augmentation or domain adaptation strategies to cope with the limited amount of labelled data.

\bibliographystyle{IEEEtran.bst}
\bibliography{IEEEabrv, biblio}

\begin{IEEEbiography}[{\includegraphics[width=1in,height=1.25in,clip,keepaspectratio]{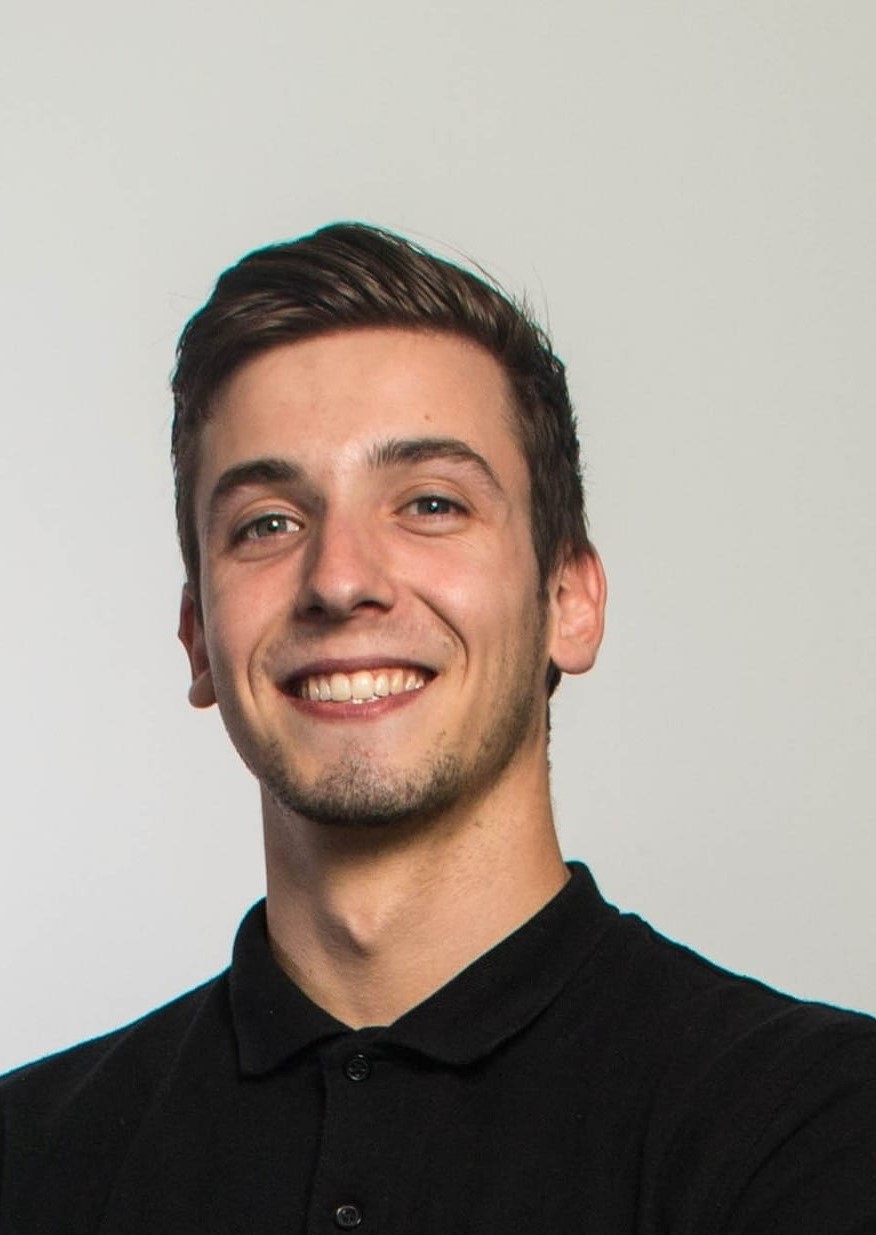}}]{Colin Decourt}
is a PhD candidate as part of ANITI (Artificial Natural Intelligence Toulouse Institute). He received his Master’s degree in telecommunications and artificial intelligence at Bordeaux Institute of Technology in 2020.
He started his PhD with ISAE-SUPAERO, CerCo (CNRS UMR5549) and NXP Semiconductors in October 2020. His research focuses on scene understanding for automotive FMCW radars (targets detection, classification and tracking) using artificial intelligence.

His research interest includes image and radar processing, computer vision, and self-supervised learning. 
\end{IEEEbiography}
\begin{IEEEbiography}[{\includegraphics[width=1in,height=1.25in,clip,keepaspectratio]{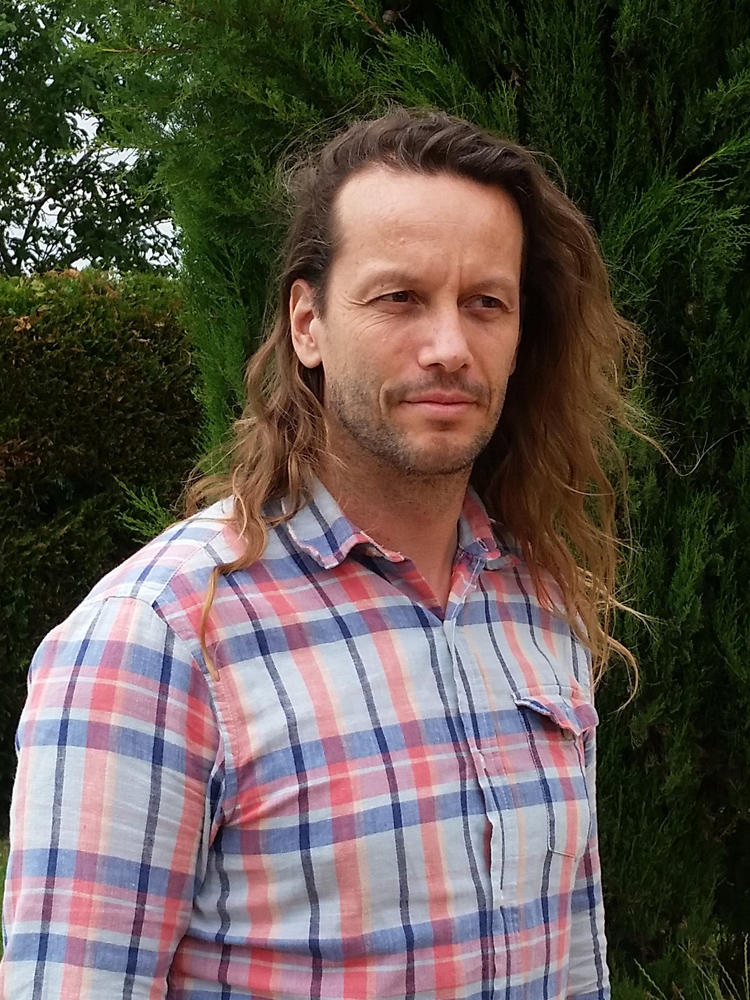}}]{Rufin VanRullen}
studied Mathematics and Computer Science, then quickly turned to Cognitive Sciences. During his PhD from Université Paul Sabatier, Toulouse, France (2000), he worked on neural coding and rapid visual processing under the supervision of Simon J. Thorpe. As a post-doctoral researcher at the California Institute of technology with Cristof Koch, he became interested in the mechanisms of visual attention and consciousness. Since 2002, he is a Research Director at the CNRS in the Brain and Cognition Research Center (CerCo) of Toulouse. His work in experimental and computational neuroscience explored the role of brain oscillations in cognition. In particular, he demonstrated that rhythmic brain activity makes our perception periodic–-a rapid sequence of perceptual cycles, akin to a video sequence. More recently, his research focuses on AI and deep neural networks. He holds a Research Chair in the Artificial and Natural Intelligence Toulouse Institute (ANITI) and has received several European grants (European Young Investigator Award, ERC Consolidator grant, ERC Advanced grant) as well as the CNRS bronze medal in 2007.
\end{IEEEbiography}
\begin{IEEEbiography}[{\includegraphics[width=1in,height=1.25in,clip,keepaspectratio]{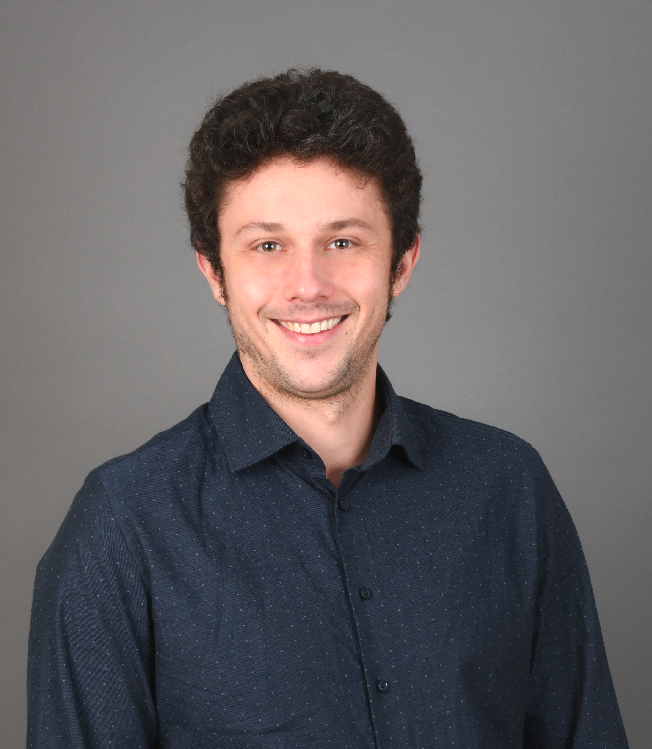}}]{Thomas Oberlin}
received the M.S. degree in applied mathematics from Université Joseph Fourier, Grenoble, France, in 2010, as well as an engineer's degree from Grenoble Institute of Technology. In 2013, he received the Ph.D. in applied mathematics from the University of Grenoble. In 2014, he was a post-doctoral fellow in signal processing and medical imaging at Inria Rennes, France, before joining as an Assistant Professor INP Toulouse -- ENSEEIHT and the IRIT Laboratory, Universit\'e de Toulouse, France. Since 2019, he is an Associate Professor in artificial intelligence and image processing at ISAE-SUPAERO, Université de Toulouse, France.  

His research interests are in signal, image and data processing and in particular time-frequency analysis, representation learning, and sparse/low-rank regularizations for inverse problems. 

Since 2022, he serves as an Associate Editor for the IEEE Transactions on Signal Processing.
\end{IEEEbiography}

\end{document}